\pdfoutput=1

\documentclass[11pt]{article}

\usepackage{ACL2023}

\usepackage{times}
\usepackage{latexsym}

\usepackage[T1]{fontenc}

\usepackage[utf8]{inputenc}

\usepackage{microtype}

\usepackage{inconsolata}

\usepackage{subcaption} 
\usepackage{adjustbox}

\usepackage{enumitem} 
\usepackage{color,soul} 

\renewcommand{\colorbox}[2]{\colorlet{tmpcolor}{#1}\sethlcolor{tmpcolor}\hl{#2}}
\usepackage{changes}
\definechangesauthor[color=yellow]{TODO}
\definechangesauthor[color=gray]{PAR} 
\definechangesauthor[color=purple!50]{BIB}
\definechangesauthor[color=purple]{REF}
\definechangesauthor[color=red]{ERR}

\newcommand{\lm}[1]{\texttt{#1}}

\usepackage{booktabs} 
\usepackage{graphicx} 
\usepackage{multirow} 
\usepackage{amsmath}
\usepackage{amsfonts}
\usepackage{amssymb}
\usepackage{bm} 
\usepackage{framed} 

\usepackage{tikz}
\usepackage{pgfplots}
\pgfplotsset{compat=1.16}

\usepackage{xcolor}

\usepackage{soul}

\definecolor{lemonchiffron}{cmyk}{0, 0.039, 0.188, 0}
\definecolor{deeppeach}{cmyk}{0, 0.203, 0.341, 0}
\definecolor{palechestnut}{cmyk}{0, 0.239, 0.226, 0.133}
\definecolor{ube}{cmyk}{0.286, 0.301, 0, 0.219}
\definecolor{crystal}{cmyk}{0.239, 0, 0, 0.149}

\definecolor{gptcolor}{rgb}{0.457, 0.66, 0.609}
\definecolor{smvcolor}{rgb}{0.6627, 0.4039, 0.3098}

\usepackage{csquotes}

\defcitealias{treviso-martins-2020-explanation}{T\&M}

\title{
    Saliency Map Verbalization: Comparing Feature Importance Representations from Model-free and Instruction-based Methods
}

\newcommand{\affilsup}[1]{\rlap{\textsuperscript{\normalfont#1}}}
\author{
    Nils Feldhus\affilsup{1}
    \qquad
    Leonhard Hennig\affilsup{1}
    \qquad
    Maximilian Dustin Nasert\affilsup{1,2}
    \\
    \textbf{Christopher Ebert}\affilsup{1,2}
    \qquad 
    \textbf{Robert Schwarzenberg}
    \qquad 
    \textbf{Sebastian M\"oller}\affilsup{1,2}
    \\
    $^1$ German Research Center for Artificial Intelligence (DFKI) \\
    $^2$ Technische Universit\"at Berlin \\
    \texttt{\{firstname.lastname\}@dfki.de} \\
}

\begin{document}
\maketitle
\begin{abstract}
Saliency maps can explain a neural model's predictions by identifying important input features. They are difficult to interpret for laypeople, especially for instances with many features.
In order to make them more accessible, we formalize the underexplored task of translating saliency maps into natural language and compare methods that address two key challenges of this approach -- what and how to verbalize. 
In both automatic and human evaluation setups, using token-level attributions from text classification tasks, we compare two novel methods (search-based and instruction-based verbalizations) against conventional feature importance representations (heatmap visualizations and extractive rationales), measuring simulatability, faithfulness, helpfulness and ease of understanding.
Instructing \lm{GPT-3.5} to generate saliency map verbalizations yields plausible explanations which include associations, abstractive summarization and commonsense reasoning, achieving by far the highest human ratings, but they are not faithfully capturing numeric information and are inconsistent in their interpretation of the task.
In comparison, our search-based, model-free verbalization approach efficiently completes templated verbalizations, is faithful by design, but falls short in helpfulness and simulatability.
Our results suggest that saliency map verbalization makes feature attribution explanations more comprehensible and less cognitively challenging to humans than conventional representations.
\footnote{Code and data at \url{https://github.com/DFKI-NLP/SMV}.}

\end{abstract}

\begin{figure*}[ht!]
    \centering
    \resizebox{\textwidth}{!}{%
        \includegraphics{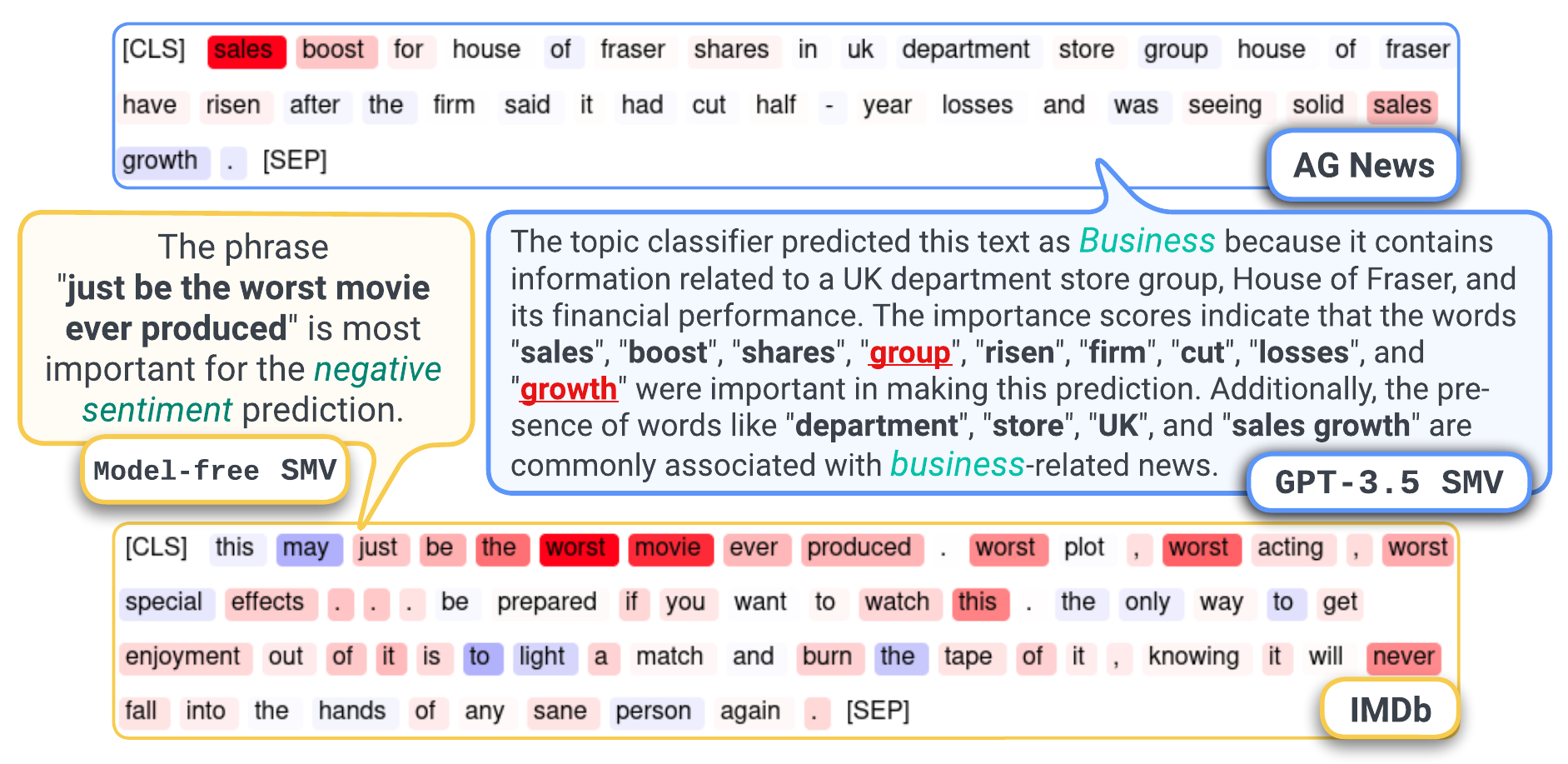}
    }
    \caption{Heatmap visualizations generated by the Integrated Gradients feature attribution method explaining the predictions of a \lm{BERT} model: Correct classifications of an instance from \textsf{AG News} (top) as \textit{Business} and an instance from \textsf{IMDb} (bottom) as \textit{Negative sentiment}. 
    Tokens with red backgrounds have higher importance scores, while blue backgrounds indicate the contrast case.
    Two verbalizations (SMVs) are depicted in the center of the figure: The left (yellow) is produced by our model-free approach, while the right (blue) is produced by \lm{GPT-3.5}.
    The predicted labels are highlighted in cyan and italic.
    The model-generated verbalization conveys semantic information such as associations with the target label (\textit{Business}) and reasoning that is disconnected from the underlying model. \lm{GPT-3.5} wrongly deems two of the least attributed tokens salient (``group'' and ``growth'', highlighted in red).}
    \label{fig:heatmap}
\end{figure*}
\section{Introduction}

Feature attribution methods, or (input) saliency methods, such as attention- or gradient-based attribution, are the most prominent class of methods for generating explanations of NLP model behavior \cite{wallace-2020-interpreting,madsen-2022-post-hoc}
and can be used to produce word-level importance scores without human supervision \cite{wallace-2019-allennlp,sarti-2023-inseq}.
A major limitation of saliency maps is that they require expert knowledge to interpret \cite{alvarez-melis-2019-weight,colin-2021-what-cannot}.
Furthermore, \citet{schuff-2022-human} revealed visual perception and belief biases which may influence the recipient's interpretation.

Natural language explanations (NLEs), on the other hand, exceed other explainability methods in plausibility \cite{lei-2016-rationalizing,wiegreffe-pinter-2019-attention,jacovi-goldberg-2020-towards}, accessibility \cite{ehsan-riedl-2020-hcxai}, and flexibility \cite{brahman-2021-learning,chen-2023-rev}, i.e.\ they can be adapted to both different target tasks and different audiences.
Most previous approaches in generating NLEs depend on datasets of human-annotated text highlights \cite{zaidan-2007-using,lei-2016-rationalizing,wiegreffe-marasovic-2021-review} or carefully constructed gold rationales for supervised training \cite{camburu-2020-make,wiegreffe-2022-reframing}, which are costly to obtain and task-specific. 
Alignment of model rationales with very few human-acceptable gold rationales may raise issues of trust \cite{jacovi-2021-formalizing-trust} and the models trained on them may suffer from hallucinations \cite{maynez-2020-faithfulness}.

In this work, we revisit and formalize the task of verbalizing saliency maps, i.e. translating the output of feature attribution methods into natural language \cite{forrest-2018-towards,mariotti-2020-towards,slack-2022-talktomodel}.
Verbalizations can describe relations between words and phrases and their associated saliency scores. Contrary to conventional heatmap visualizations, we can adjust the comprehensiveness of an explanation more precisely and infuse it with additional semantics such as word meanings, concepts, and context about the task.

We find that verbalization also comes with a few caveats:
Similar to human explainers, who communicate only the most relevant explanations to avoid cognitive overload of the recipient \cite{hilton-2017-social,miller-2019-explanation},
verbalization methods need to address the problem of deciding ``what'' to say, i.e. selecting the most informative and useful aspects of the saliency maps and communicating them in a concise manner.
We therefore compare different methods for verbalizing saliency maps: Supervised rationales, prompting LLMs, and model/training-free templates.

We address the problem of saliency map verbalization (\textbf{SMV}) with the following contributions:
\begin{itemize}[noitemsep,topsep=0pt,leftmargin=*]
    \item We formalize the underexplored task of SMV and establish desiderata, i.e. simulatability, explainer-faithfulness, plausibility, and conciseness (\S \ref{sec:formalization});

    \item We conduct a comparative study on various representations of feature attribution in two text classification setups, measuring the effects of verbalizations methods on both automated (
    explainer-faithfulness) and human evaluation metrics (simulatability, helpfulness, ease of understanding) (\S \ref{sec:study}, \S \ref{sec:experiments}).

    \item We propose a novel, model-free, template-based SMV approach, and design instructions for \lm{GPT-3.5}-generated SMVs (\S \ref{sec:method}) (examples from our two setups are depicted in Fig.~\ref{fig:heatmap});

    \item We show that model-free SMVs perform slightly better than heatmaps and extractive rationales on ease of understanding and are faithful by design, while instruction-based SMVs achieve the highest average simulation accuracy and are preferred in subjective ratings (\S \ref{sec:results});

    \item We publish a large dataset of model-free and \lm{GPT}-generated SMVs alongside extractive rationales and results from both evaluations, and open-source code to produce all kinds of SMVs.
    
\end{itemize}
 
\section{Verbalizing saliency maps}

\subsection{Formalization}
\label{sec:formalization}

The setup of the saliency map verbalization task consists of 
an underlying (to-be-explained) \textbf{model}~$m$
whose prediction $\hat{y}~\subset~Y$ on source tokens $W=w_1~\dots~w_n$ we want to explain (against the set of possible outcomes $Y$).

$m$ is equipped with
a feature explanation method (or short: \textbf{explainer}) $e$
which produces a \textbf{saliency~map} $S=s_1 \dots s_n$:

\vspace*{-10pt}
\begin{equation}
    e(W, m) = S
\end{equation}

Here, we call token~$w_i$ salient \textit{towards} outcome~$y$
if its associated saliency score $s_i > 0$
and salient \textit{against} $y$ for $s_i < 0$. 
$e$ can have many sources, e.g. gradient-based methods such as Integrated Gradients \cite{sundararajan-2017-axiomatic} which we employ in our experiments (\S \ref{sec:experiments}), or even human experts assigning relevance scores.

A \textbf{verbalized saliency map} $S_{\text{V}}$ is produced by some verbalizer $v$ that receives the output of $e$:

\vspace*{-15pt}
\begin{equation}
\label{eqn:vex}
    v(W, S) = S_{\text{V}}
\end{equation}

$v$ can be any function that discretizes attribution scores and constructs a natural language representation $S_{\text{V}}$.
This is connected to the concept of hard selection in \citet{deyoung-2020-eraser} and heuristics for discretizing rationales \cite{jain-2020-fresh}.
In the taxonomy of \citet{wiegreffe-marasovic-2021-review},
verbalized saliency maps can be categorized as free-text rationales with varying degrees of structure imposed through templates.
Moreover, verbalized explanations are procedural and deterministic by nature,
i.e. they function as instructions that one can directly follow \cite{tan-2022-diversity}
to understand a model's decision,
similar to compositional explanations \cite{hancock-2018-babblelabble,yao-2021-refining}.

\subsection{Desiderata}
\label{sec:desiderata}

In the following, we outline the common evaluation paradigms for explanations (faithfulness, simulatability, plausibility) and how we adapt them to saliency map verbalizations.

\paragraph{Faithfulness}
Saliency maps express that ``certain parts of the input are more important to the model reasoning than others'' (\textit{linearity assumption} in \citet{jacovi-goldberg-2020-towards}).
For verbalizations, explainer $e$ and verbalizer $v$ are two separate processes, so the saliency map $S$ can be seen as static.
Therefore, the faithfulness of $e$ to the model~$m$ is extrinsic to the verbalization.
Instead, it is essential to faithfully translate $S$ into natural language, which we coin \textbf{explainer-faithfulness}. The verbalizer breaks faithfulness, e.g. if words are referenced as salient in $S_{\text{V}}$ that are made up (do not appear in $W$) or if the polarity of any $s_i$ is falsely interpreted.

\paragraph{Simulatability} 
Another type of faithfulness is the model assumption
which requires two models to ``make the same predictions [iff] they use the same reasoning process'' \cite{jacovi-goldberg-2020-towards}.
By extension this means a model has to be simulatable \cite{doshi-velez-kim-2017-rigorous,hase-bansal-2020-evaluating},
i.e. a human or another model should be able to predict a model's behaviour on unseen examples while exposed only to the explanation and not the model's prediction.

\paragraph{Plausibility}
The plausibility of explanations is commonly measured by correlation with ground-truth explanations \cite{deyoung-2020-eraser,jacovi-goldberg-2020-towards},
since gold rationales are influenced by human priors on what a model should do.

\paragraph{Conciseness}
In addition to these paradigms, verbosity is also an important aspect.
A full translation into natural language is nonsensical, however, because all relations between the continuous-valued saliency scores and the associated tokens would normally overload human cognitive abilities.
We want $S_{\text{V}}$ to be concise, yet still contain the key information, similar to sufficiency and comprehensiveness measures from \citet{deyoung-2020-eraser}.
Thus, we define a \textbf{coverage} measure to indicate how much information is retained going from $S$ to $S_{\text{V}}$, i.e. how much of the total attribution in $S = s_1 \dots s_n$ is referenced by the tokens mentioned in $S_{\text{V}} = v_1 \dots v_m$:

\vspace*{-10pt}
\begin{equation}
\label{eqn:cov}
    \text{Coverage}(S_{\text{V}}) = \frac{\sum |v_i|}{||S||}
\end{equation}

The goal here is not to achieve a coverage of $1$ with all of $S$, but depending on the use case, $S_{\text{V}}$ should mention the most influential tokens, so a trivial solution for $k=5$ would be to include the top $k$ tokens with the highest attribution in $S$.

\section{Study setup}
\label{sec:study}

\subsection{Human Evaluation}
\label{sec:human_eval}

Inspired by previous crowd studies in explainability \cite{chandrasekaran-2018-vqa,strout-2019-human-rationales,hase-bansal-2020-evaluating,sen-2020-human-attention,gonzalez-2021-interaction,arora-2022-explain-edit-understand,joshi-2023-machine-rationales}, we propose to measure \textbf{simulatability} as well as ratings for helpfulness and ease of understanding (\textbf{plausibility}). We evaluate the quality of different verbalization methods in a study involving 10 human participants. All participants have a computational linguistics background, with at least a Bachelor's degree, limited to no prior exposure to explainability methods, and are proficient in English (non-native speakers). After an introduction to the goal of the study and a brief tutorial, annotators are to complete the tasks described below. For each task, we present text instances along with their explanations, using a simple Excel interface.\footnote{See Appendix~\ref{app:efficiency}, Figure~\ref{fig:interface}}

\paragraph{Task A: Simulation}
In the first task, participants are asked to simulate the model, i.e.\ predict the model's outcome, based only on one type of explanation plus the input text (``What does the model predict?''). They are given the possible class labels and were given an example for each dataset in the tutorial before starting the session. If the explanation does not provide any sensible clues about the predicted label, they still have to select a label, but may indicate this in the following question B1.

\paragraph{Task B: Rating}
In the second task, participants have to provide a rating on a seven-point Likert scale about (B1) ``how helpful they found the explanation for guessing the model prediction'' and (B2) ``how easy they found the explanation to understand''. A higher rating indicates a higher quality of the explanation.

\paragraph{Task C: Questionnaire}
Finally, participants are asked to complete a post-annotation questionnaire to obtain overall judgements for each verbalization method. They are prompted for Likert scale ratings about time consumption, coherence, consistency and qualitative aspects of each verbalization method, as listed in Table~\ref{tab:quest}.

\subsection{Automated Evaluation}
\label{sec:autoeval}

We expect hallucinations (synthesized, factually incorrect text due to learned patterns and statistical cues) from \lm{GPT}-type models and thus devise the following tests measuring \textbf{explainer-faithfulness} and \textbf{conciseness}:

\begin{enumerate}[noitemsep,topsep=0pt,leftmargin=*]
    \item Have the referred words been accurately cited from the input text?
    \item How often do the referred words represent the top $k$ most important tokens? (Eq.~\ref{eqn:cov})
\end{enumerate}

We obtain the results by simple counting and automated set intersection.

\begin{figure*}[ht!]
    \centering
    \resizebox{0.975\textwidth}{!}{%
        \includegraphics{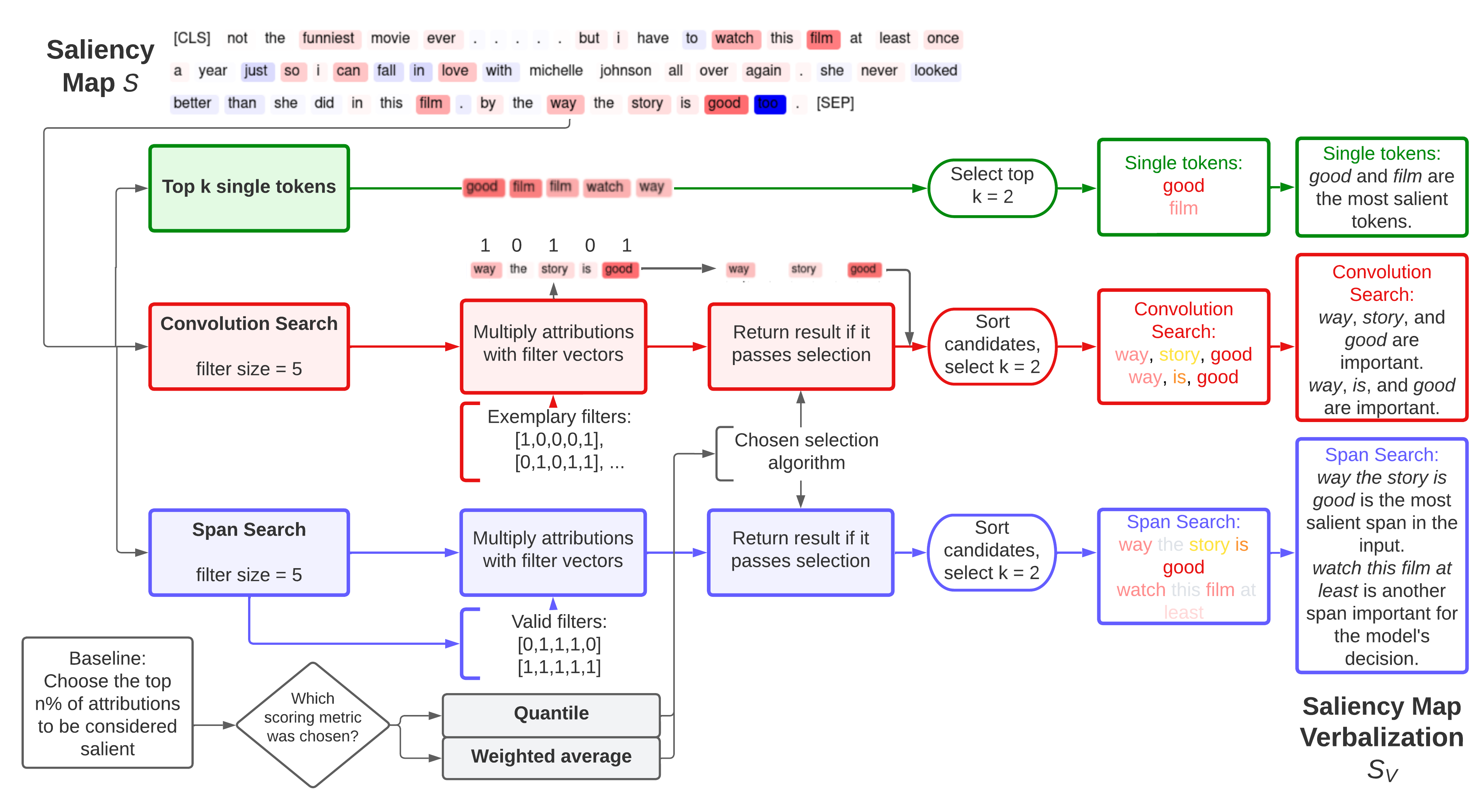}
    }
    \caption{Model-free saliency map verbalizations (SMV$_{\text{Templ}}$) as generated from three different search methods (Top $k$ single tokens, Convolution Search, Span Search) and two scoring metrics (Quantile, Weighted Average).}
    \label{fig:method}
\end{figure*}

\section{Methods}
\label{sec:method}

To complement heatmap visualizations and extractive rationales, we propose and analyze two additional verbalization methods: Model-free (\S \ref{sec:modelfree}, Fig.~\ref{fig:method}) and instruction-based (\S \ref{sec:gpt}, Fig.~\ref{fig:gptmethod}) saliency map verbalization.

\subsection{Model-free verbalization}
\label{sec:modelfree}

For our model-free approach we employ hand-crafted templates for surface realization, different binary filter algorithms as search methods (\S \ref{sec:search}) and scoring metrics (\S \ref{sec:scoring}) to select tokens for filling the templates.
This approach does not require architectural changes to the underlying model or modifications to an existing saliency method.
The most similar approach to our selection heuristics, to our knowledge, are the discretization strategies in \citet[\S 5.2]{jain-2020-fresh}. 

In the following, we will present two distinct candidate generation methods that can both be combined with one of two scoring metrics. A final candidate selection (\S \ref{sec:compare}) will collect the results from both searches, concatenate them to possibly larger spans and filter the top scoring candidates once more while maximizing coverage (Eq. \ref{eqn:cov}). These salient subsets are then used to complete hand-crafted templates (App.~\ref{app:templates}).
We argue that this is more human-interpretable than simple top $k$ single token selection, at the cost of a lower coverage. Our methodology allows to set parameters in accordance to how faithful the verbalization should be to the underlying explainer.

\subsubsection{Explanation search}
\label{sec:search}

To acquire potentially salient snippets from a given text, we perform a binary selection on a window of attributions from the input of size $c$ and then compare the sum of our selection to one of our scoring methods, performing basic statistical analysis on the window and the input.

\paragraph{Convolution Search}
Inspired by the convolutions of neural networks, we compare tokens that are located close to each other but are not necessarily direct neighbors. Coherence between pairs of tokens is solely determined by looking at their attributions with the following binary filters.
In short, the following method firstly generates template-vectors that we then permute and keep as our binary filters. After computing all valid and sensible permutations, we can start calculating possibly salient or coherent snippets of our input.
We choose $b\in \mathbb{N}$ vectors with a length of $c \in \mathbb{N}$.
We describe these $b$ vectors $v_i$ as follows:

\vspace*{-10pt}
\begin{equation}
    v_i = [1_{1,i}, 0_{1, c-i}],\ \ v_i\in \mathbb{Z}^{1,c}. \\
\end{equation}
\vspace*{-30pt}
\begin{gather*}
\text{e.g., for $i=3$, $c=5$, $v_i=$}
    \begin{smallmatrix}
    (1 & 1 & 1 & 0 & 0)
    \end{smallmatrix}
\end{gather*}
\vspace*{-20pt}

We only keep those $v_i$ where $\sum v_i \notin \{0,1,c\}$ in order to perform sensible permutations. 
For each $v_i$, we define a filter $\bm{f}_{i,j}$, where each distinct entry in $\bm{f}_i$ is a unique permutation of $v_i$.
Let $A$ be our attribution input, with $A\in\mathbb{R}^{1,k}$, where $k$ is the length of our input $k>c$, then we multiply a subset of our input with every binary filter 

\vspace*{-20pt}
\begin{equation}
    \begin{split}
    \bm{r}_{i,j,l} = \bm{f}_{i,j} \cdot A_{l}^{l+c}, \\ l\in L, 
    L=\{l\in\mathbb{Z}|1\leq l\leq k-c\}.
    \end{split}
\end{equation}

\vspace*{-5pt}
From this, we receive result vectors containing possibly coherent attributions and tokens.

\paragraph{Span Search}
Instead of looking for token pairs in a local neighborhood, we can also look for contiguous spans of tokens by adapting our proposed convolutional search.\\
We generate $b$ vectors of length of $c$ with $c$ being odd.
We describe these $b$ vectors as follows: Choose $i \in \mathbb{N}$ with $i$ being odd, which ensures symmetry of our filters.\footnote{In contrast to our proposed Convolution Search, we don't need permutations of $v_i$ to generate filters $\mathbf{f}$, so we directly use $v_i$. Thus, the result vector $\mathbf{r}$ has only two indices.}
\begin{equation}
    v_i = [0_{1, \lfloor\frac{c-i}{2}\rfloor}, 1_{1,i}, 0_{1, \lfloor\frac{c-i}{2}\rfloor}],\ \ v_i\in \mathbb{Z}^{1,c}
\end{equation}
We calculate attribution vectors $\bm{r}_{i,l}$ as such:
\begin{equation}
    \begin{split}
    \bm{r}_{i,l} = v_{i} \cdot A_{l}^{l+c}, \\
    l\in L, L=\{l\in\mathbb{Z}|1\leq l\leq k-c\}
    \end{split}
\end{equation}

\subsubsection{Candidate scoring metrics}
\label{sec:scoring}
We score and filter the snippets $\mathbf{r}$ so that we can present the most salient samples. As a threshold, we calculate the average of the $n\%$ most salient tokens of the given input sample $A$. This simple method does not filter for saliency, but it reduces the likelihood of presenting non-salient sample snippets. We call this our baseline $\beta$.

\paragraph{Weighted average}
The weighted average sums up the attribution values of $r$ and divides the resulting scalar by the length of $r$, calculating the "saliency per word" of $r$. Then the result gets compared to $\beta$. Is the result larger than $\beta$, $r$ is considered salient and will be a candidate for the verbalization.

\paragraph{Quantile}
The quantile method relies on the standard deviation within our current sample $A$. Given a quantile $n, n \in \mathbb{R}^+_0$, we calculate the corresponding standard deviation value $\sigma$ and compare it to the average of the values of our snippet. If the score is greater than $\sigma$ and $\beta$, it will be marked for verbalization.

\subsubsection{Summarized explanation}
\label{sec:compare}

On top of the two search methods in \S \ref{sec:search}, we construct a summarized explanation to be used in our human evaluation (\S \ref{sec:human_eval}) by considering the $k$ single tokens with the highest attribution scores.
After generating $k$ candidates from each search method, we concatenate neighboring token indices to (possibly) longer sequences and recalculate their coverage.
We compute the $q$-th quantile of the remaining candidates according to their coverage to select the final input(s) to our templates. If no candidate is within the $q$-th quantile, the top-scoring span will be chosen.

\begin{figure}[t!]
    \centering
    \resizebox{\columnwidth}{!}{%
    \includegraphics{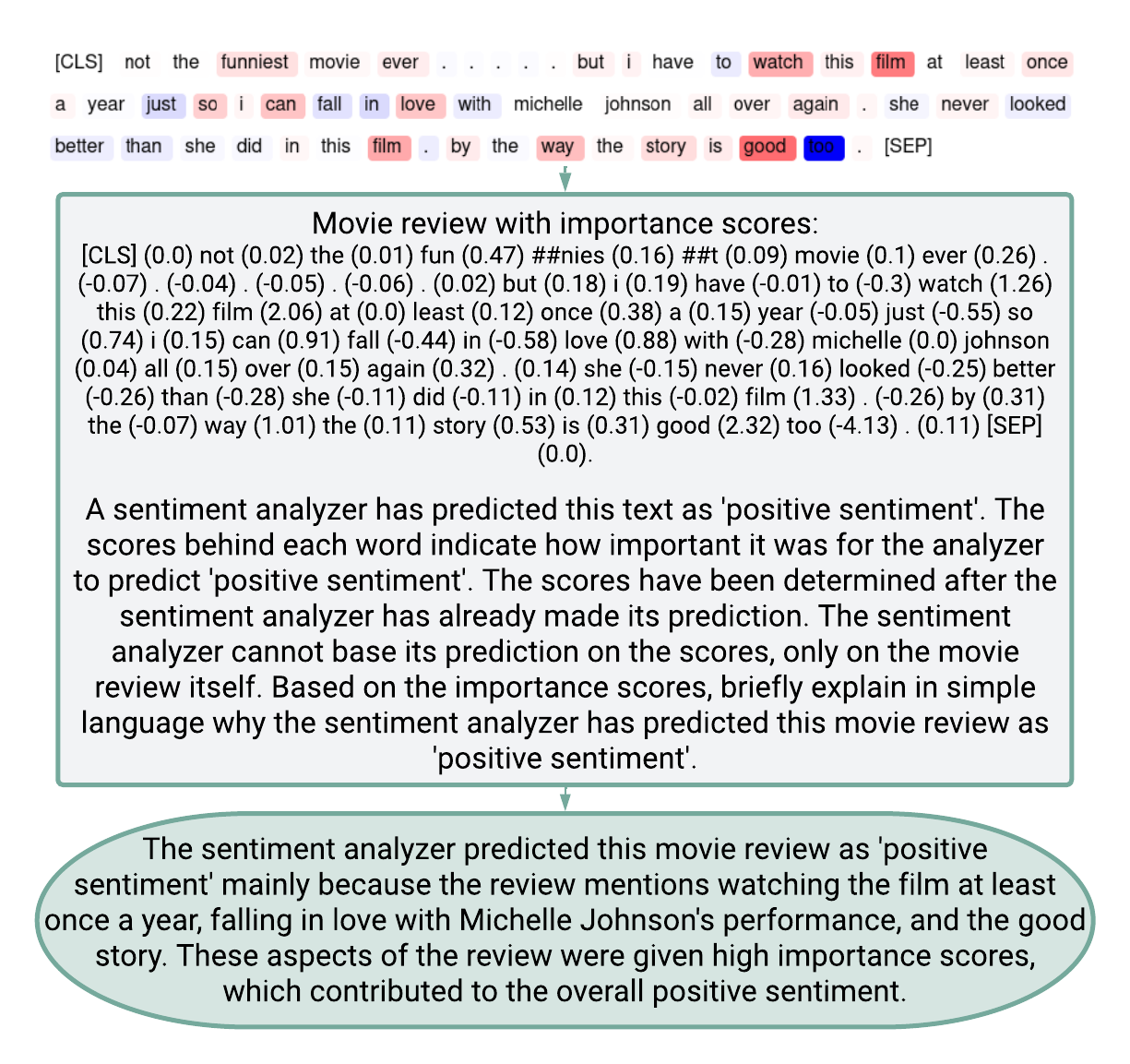}}
    \caption{Instruction-based verbalizations SMV$_{\texttt{GPT}}$ using \lm{GPT-3.5} of a \textit{negative sentiment} instance from \textsf{IMDb} that was wrongly classified by \lm{BERT}.}
    \label{fig:gptmethod}
\end{figure}

\subsection{Instruction-based Verbalizations}
\label{sec:gpt}
In light of very recent advances in instructing large language models to perform increasingly complex tasks \cite{wei-2022-finetuned}, we additionally construct ``rationale-augmented'' verbalizations (Fig.~\ref{fig:gptmethod}) next to template-based and search-based ones. The instruction contains an overview of the saliency map verbalization task and the associated caveats, e.g. ``The classifier cannot base its prediction on the scores, only on the input text itself.''. Our most consistently accurate result was achieved by then representing $S$ as bracketed scores rounded to two digits put behind each word, e.g. ``definitely (0.75) a (0.14) girl (-0.31) movie (0.15)''.

In practice, we manually engineered task-agnostic instruction templates to work with \lm{GPT-3.5} (March '23) aka \lm{ChatGPT}.\footnote{We describe the task-specific instructions in App.~\ref{app:prompts} and document the edits to mitigate label leakage in App.~\ref{app:gpt-postproc}.}
To our knowledge, there are no datasets with gold verbalizations available and we do not want to enforce any specific format of the explanation, so we use the API in a zero-shot setting. We post-process all outputs by removing all occurrences of the predicted label and semantically very similar words (App.~\ref{app:gpt-postproc}). 

\begin{table}[t!]
    \centering
    
    \resizebox{\columnwidth}{!}{%
        \renewcommand{\arraystretch}{0.85}
    
        \begin{tabular}{r|ll}
            \toprule
            \textit{Explanations...}
            & \textbf{Templ}
            & \textbf{\texttt{GPT}} \\
            
            \midrule 
            were concise \& not time-consuming.
                & 4.00
                & 2.38
                \\
            were not too complex.
                & 3.63
                & 3.88
                \\
            were not inconsistent/contradictory.
                & -
                & 3
                \\
            helped me detect wrong predictions.
                & 2.63
                & 3
                \\
            with more diverse sentences are useful.
                & 4.25*
                & -
                \\
            with numeric scores are useful.
                & 2.63*
                & 2.38
                \\
            with associations/context are useful.
                & 4.00*
                & 4.50
                \\ 
            summarizing the input are useful.
                & -
                & 4.75
                \\
        
            \bottomrule
        \end{tabular}
    }
    \caption{Questionnaire asking participants about their overall impressions on both types of verbalizations. All aspects were rated based on a 5-point Likert scale (1: ``strongly disagree''; 5: ``strongly agree''). Starred values: SMV$_{\text{Templ}}$ do not have this property, so we asked if the participants \textit{would have liked them} to have it.
    }
    \label{tab:quest}
\end{table}

\section{Data}
\label{sec:experiments}

We choose datasets that cover a selection of English-language text classification tasks.
In particular, we select \textsf{IMDb} \cite{maas-2011-imdb} for sentiment analysis, and \textsf{AG News} \cite{zhang-2015-character} for topic classification.

We retrieve predictions from \lm{BERT} models on the test partitions of \textsf{IMDb} and \textsf{AG News} made available through TextAttack \cite{morris-2020-textattack} and their Integrated Gradients \cite{sundararajan-2017-axiomatic} explanations with 25 samples exactly as they appear in Thermostat \cite{feldhus-2021-thermostat}.

\begin{table*}[ht!]
    \centering
    \resizebox{\textwidth}{!}{%
        \renewcommand{\arraystretch}{0.95}
        \begin{tabular}{cr|cccc|cccc|cccc|}
        \cmidrule[0.5pt](l{0.5em}r{0.5em}){3-14}
        &
        & \multicolumn{4}{c|}{\textbf{A: Simulation Accuracy}} 
        & \multicolumn{4}{c|}{\textbf{B1: Helpfulness}} 
        & \multicolumn{4}{c|}{\textbf{B2: Ease of understanding}} \\
        \cmidrule[0.25pt](l{0.25em}r{0.25em}){3-14}
        
        &
            & $\underset{\text{Vis}}{\text{HM}}$ 
            & $\underset{\text{Extr}}{\text{Rat}}$ 
            & $\underset{\text{Templ}}{\text{SMV}}$ 
            & $\underset{\texttt{GPT}}{\text{SMV}}$
            & $\underset{\text{Vis}}{\text{HM}}$ 
            & $\underset{\text{Extr}}{\text{Rat}}$ 
            & $\underset{\text{Templ}}{\text{SMV}}$ 
            & $\underset{\texttt{GPT}}{\text{SMV}}$
            & $\underset{\text{Vis}}{\text{HM}}$ 
            & $\underset{\text{Extr}}{\text{Rat}}$ 
            & $\underset{\text{Templ}}{\text{SMV}}$ 
            & $\underset{\texttt{GPT}}{\text{SMV}}$ \\
        
        \cmidrule[1pt](l{0.5em}r{0.5em}){2-14}
        & All
            & 90.75 & 85.94 & 87.5 & \textbf{94.06}
            & 4.73 & 4.19 & 4.46 & \textbf{5.80}
            & 4.35 & 4.00 & 4.67 & \textbf{5.88} \\
        \textsf{IMDb} 
        & \colorbox{lemonchiffron}{$\text{Cov}(\text{S}_{\text{VT}})^{\nearrow}$}
            & 94.38 & 89.45 & 92.19 & \textbf{96.09}
            & 4.98 & 4.50 & 4.91 & \textbf{5.94}
            & 4.47 & 4.34 & 4.99 & \textbf{5.99}
            \\
        {$_{\text{IAA}}$}
        & \colorbox{crystal}{$y \neq \hat{y}$} 
            & 74.49 & 58.43 & 63.90 & \textbf{84.65} 
            & 3.67 & 3.09 & 3.21 & \textbf{5.01}
            & 3.48 & 2.92 & 3.61 & \textbf{5.25} \\
        {$\kappa = 0.731$}
        & \colorbox{deeppeach}{$\hat{y} \neq y_{\text{sim}}$}
            & \multicolumn{4}{c|}{n.a. (0.00)}
            & 3.40 & 3.10 & 2.85 & \textbf{3.94} 
            & 3.48 & 3.18 & 3.35 & \textbf{4.33}
            \\
    
        \cmidrule[0.5pt](l{0.5em}r{0.5em}){2-14}
        & All
            & \textbf{79.83} & - & 79.50 & 77.60
            & 5.26 & - & 4.65 & \textbf{5.63}
            & 5.02 & - & 4.90 & \textbf{5.77}
            \\
        \textsf{AG News}
        & \colorbox{lemonchiffron}{$\text{Cov}(\text{S}_{\text{VT}})^{\nearrow}$}
            & \textbf{85.31} & - & 84.57 & 81.13
            & 5.41 & - & 4.98 & \textbf{5.80}
            & 5.18 & - & 5.13 & \textbf{5.89}
            \\
        {$_{\text{IAA}}$}
        & \colorbox{crystal}{$y \neq \hat{y}$} 
            & \textbf{70.17} & - & 69.37 & 64.53
            & 5.02 & - & 4.52 & \textbf{5.36}
            & 4.84 & - & 4.84 & \textbf{5.61} 
            \\
        {$\kappa = 0.721$}
        & \colorbox{deeppeach}{$\hat{y} \neq y_{\text{sim}}$}
            & \multicolumn{4}{c|}{n.a. (0.00)}
            & 4.14 & - & 3.34 & \textbf{4.40}
            & 4.08 & - & 3.89 & \textbf{5.10}
            \\
        
        \cmidrule[1pt](l{0.5em}r{0.5em}){2-14}
        \end{tabular}
    }
    
    \caption{Results of the human evaluation. Task \textbf{A}: Simulation accuracy (annotators guessing the label predicted by the underlying \lm{BERT} correctly). Task B: Average rating of annotators (1 ``bad'' - 7 ``good'') for helpfulness (\textbf{B1}) and ease of understanding (\textbf{B2}). 
    HM-Vis = Heatmap visualization. Rat-Extr = Extractive rationalizer of \citet{treviso-martins-2020-explanation}. SMV-Templ = Template-based saliency map verbalization. SMV-\lm{GPT} = \lm{GPT-3.5}-based saliency map verbalization.
    \textbf{All}: Overall result. 
    \textbf{$\text{Cov}(\text{S}_{\text{VT}})^{\nearrow}$}: Coverage above average.
    \textbf{$y \neq \hat{y}$}: Explained \lm{BERT} model made a false prediction. 
    \textbf{$\hat{y} \neq y_{\text{sim}}$}: False human simulation.
    Inter-annotator agreement in Fleiss $\kappa$ below the dataset names.
    }
    \label{tab:results_humeval}
\end{table*}

We then take subsets (\textsf{IMDb}: $n=80$, \textsf{AG News}: $n=120$) of each dataset according to multiple heuristics (App.~\ref{app:heuristics}) that make the tasks more manageable for annotators.
Each annotator was shown 340 explanations consisting of equal amounts of each type of representation or rationale.
We randomize the order in which they are presented to the annotators.
Every instance was evaluated by seven different annotators.

\section{Results}
\label{sec:results}

\paragraph{Human evaluation}
\label{sec:questionnaire}

Tab.~\ref{tab:results_humeval} shows that both kinds of SMVs are generally \textbf{easier to understand} (B2) than heatmaps or extractive rationales.
In a post-annotation questionnaire, we asked 8 out of 10 participants 14 questions about both types of SMVs. Tab.~\ref{tab:quest} lists the results. While template-based explanations are preferred in being less time-consuming, we can see that GPT-generated verbalizations outperform them in all other aspects. Unsurprisingly, associations and summarizations are the preferred characteristics of verbalizations.

\paragraph{Downstream tasks}
According to \citet{jacovi-2023-diagnosing}, a feature attribution explanation aggregates counterfactual contexts. This becomes apparent in our overall results on the \textsf{AG News} dataset where more than one potential alternative (multi-class classification with $|C|=4$) outcome exists. Annotators' simulation accuracy drops from as high as 94 \% (\textsf{IMDb}) to 78 \%. SMV$_{\texttt{GPT}}$ beats all other representations across all three measures in \textsf{IMDb}, but surprisingly underperforms in \textsf{AG News}. 

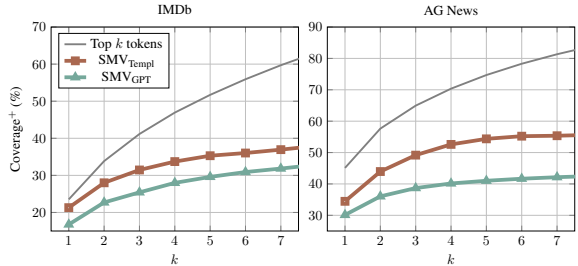
\begin{figure}[t!]
    \centering
    \begin{subfigure}[b]{0.45\columnwidth}
        \begin{tikzpicture}[scale=0.475]
        \centering
        \begin{axis}[
          title={IMDb},
          xlabel={$k$},
          ylabel={Coverage$^+$ (\%)},
          xmin=0.5, xmax=7.5,
          ymin=15, ymax=70,
          xtick={1, 2, 3, 4, 5, 6, 7, 8, 9, 10},
          ytick={20, 30, 40, 50, 60, 70},
          legend pos=north west,
          grid=major,
          ymajorgrids=true,
          xmajorgrids=true,
        ]

        \addplot[
          color=gray,
          line width=1.5pt,
          ] coordinates {
          (1, 23.53)
          (2, 33.85)
          (3, 41.16)
          (4, 46.93)
          (5, 51.71)
          (6, 55.92)
          (7, 59.70)
          (8, 63.11)
          (9, 66.25)
          (10, 69.07)
        };
        \addlegendentry{Top $k$ tokens}
        
        \addplot[
          color=smvcolor,
          line width=3pt,
          mark=square,
          ] coordinates {
          (1, 21.30)
          (2, 28.00)
          (3, 31.46)
          (4, 33.72)
          (5, 35.30)
          (6, 36.03)
          (7, 36.94)
          (8, 37.94)
          (9, 38.70)
          (10, 39.13)
        };
        \addlegendentry{SMV$_{\text{Templ}}$}
        
        \addplot[
          color=gptcolor,
          line width=3pt,
          mark=triangle,
          ] coordinates {
          (1, 16.76)
          (2, 22.70)
          (3, 25.42)
          (4, 27.98)
          (5, 29.59)
          (6, 30.90)
          (7, 31.84)
          (8, 32.86)
          (9, 33.68)
          (10, 34.43)
        };
        \addlegendentry{SMV$_{\text{GPT}}$}
        
        \end{axis}
        \end{tikzpicture}
    \end{subfigure}
    \hfill
    \begin{subfigure}[b]{0.45\columnwidth}
        \begin{tikzpicture}[scale=0.475]
        \centering
        \begin{axis}[
          title={AG News},
          xlabel={$k$},
          ylabel={},
          xmin=0.5, xmax=7.5,
          ymin=25, ymax=90,
          xtick={1, 2, 3, 4, 5, 6, 7, 8, 9, 10},
          ytick={30, 40, 50, 60, 70, 80, 90},
          grid=major,
          ymajorgrids=true,
          xmajorgrids=true,
        ]

        \addplot[
          color=gray,
          line width=1.5pt,
          ] coordinates {
          (1, 45.11)
          (2, 57.63)
          (3, 64.95)
          (4, 70.36)
          (5, 74.70)
          (6, 78.30)
          (7, 81.33)
          (8, 83.97)
          (9, 86.25)
          (10, 88.17)
        };
        \addlegendentry{Top $k$ tokens}
        
        \addplot[
          color=smvcolor,
          line width=3pt,
          mark=square,
          ] coordinates {
          (1, 34.46)
          (2, 43.95)
          (3, 49.18)
          (4, 52.60)
          (5, 54.37)
          (6, 55.22)
          (7, 55.36)
          (8, 55.65)
          (9, 55.78)
          (10, 55.97)
        };
        \addlegendentry{SMV$_{\text{Templ}}$}
        
        \addplot[
          color=gptcolor,
          line width=3pt,
          mark=triangle,
          ] coordinates {
          (1, 30.12)
          (2, 36.04)
          (3, 38.68)
          (4, 40.15)
          (5, 41.01)
          (6, 41.66)
          (7, 42.15)
          (8, 42.52)
          (9, 42.88)
          (10, 43.18)
        };
        \addlegendentry{SMV$_{\texttt{GPT}}$}

        \legend{}
        
        \end{axis}
        \end{tikzpicture}
    \end{subfigure}
    \hspace{0.025\columnwidth}
    \caption{Coverage$^{+}@k$ of SMV$_{\text{Templ}}$ and SMV$_{\texttt{GPT}}$. Top $k$ tokens is the upper bound for explainer-faithfulness.}
    \label{fig:coverage}
\end{figure}

\paragraph{Coverage of the verbalization}
Fig.~\ref{fig:coverage} and App.~\ref{app:tokenranks} show that SMV$_{\texttt{GPT}}$ focuses less on the actual most important tokens that might not be intuitive for recipients, such as function words.
The subset of instances with higher-than-average coverage according to SMV$_{\text{Templ}}$ ({\colorbox{lemonchiffron}{$\text{Cov}(\text{S}_{\text{VT}})^{\nearrow}$}}) is substantially easier to simulate (\textsf{IMDb}) and elicits the highest ratings and accuracies from annotators.
We utilize this as a proxy for (low) complexity of $S$, because usually only a single or few tokens that are very salient make these explanations easy to decipher in most representations.

Therefore, we conducted an automated simulatability evaluation on all SMV types, documented in Appendix~\ref{app:autosim}, confirming the suspicions about the faithfulness of GPT verbalizations.

\paragraph{Model predictions}
Lastly, we investigate the subsets of wrong model predictions: 
The drop in simulation accuracy and ratings when we filter the instances where the model predicts something different from the true label ({\colorbox{crystal}{$y \neq \hat{y}$}}) is more severe for \textsf{IMDb} throughout all types of explanations. In \textsf{AG News}, the simulatability and the ease of understanding turn out to be higher for SMVs. Our consistently worse results in this subset reveal the belief bias \cite{gonzalez-2021-interaction}, i.e. explanations have a hard time convincing humans about a model behavior when they already have prior assumptions about the true label of an instance.
For instances where the human simulation mismatched with the predicted label ({\colorbox{deeppeach}{$\hat{y} \neq y_{\text{sim}}$}}), the drop in scores is even harsher: Only SMV$_{\texttt{GPT}}$ still achieves ratings that are slightly above average.

\subsection{Evaluating instruction-based verbalizations}
\label{sec:gpt-eval}

While there are no invented words in the human evaluation subset, our automated mapping between explanation and input text still detected cases where words are \textbf{auto-corrected} and not accurately copied, especially fixing capitalization and small typos. We also found examples in which words or spans are replaced with a synonym, e.g. ``not reliable'' $\rightarrow$ ``unreliable'', but most strikingly, in an \textsf{IMDb} example, ``good premise'' was replaced with ``bad premise'' which entirely changed the meaning and the polarity of the sentiment. 

In Tab.~\ref{tab:gptquality}, we manually count what \textbf{type of task-related information and semantics} SMV$_{\texttt{GPT}}$ provides on top of the translation of the importance scores. 
We can see that the ``negative sentiment'' in \textsf{IMDb} is often a confounder for the correct interpretation of the negative saliency scores. 
Without explicit instructions, GPT still questioned some of the wrong prediction the underlying \lm{BERT} has made, particularly for \textsf{IMDb}.
In terms of linguistic aspects of the verbalizations, associations are frequently included, while summarizations of the input or the decision are rare.

\subsection{Discussion}
\label{sec:discussion}

By choosing parameters that prefer longer spans to be selected, we show that SMV$_{\text{Templ}}$ can be more plausible to humans than single token selection.
We acknowledge that SMV$_{\text{Templ}}$ are repetitive and, while the results show that they can guarantee a minimum degree of understandability \cite{ehsan-2019-automated-rg}, sufficiency and conciseness, they will not be satisfying enough for lay recipients on their own.

For SMV$_{\texttt{GPT}}$, 
the choice of instruction can greatly impact the faithfulness to the explainer. Plausible explanations driven by world knowledge and semantics allow laypeople to contextualize the prediction w.r.t. the input text, but reliable and generalizable methods for auditing these rationales for faithfulness have yet to be discovered.

\section{Related Work}

To our knowledge, the only previous saliency map \textbf{verbalization} approach is by \citet{forrest-2018-towards}
who used LIME explanations and a template-based NLG pipeline on a credit dataset.
While they mostly included numerical values in explanations,
we focus on most important features and free-text rationales,
because humans are more interested in reasoning than in numerical values \cite{reiter-2019-natural}.
\citet{ampomah-2022-textual-explanations} created a dataset of tables summarizing the performance metrics of a text classifier and trained a neural module to automatically generate accompanying texts.
The HCI community highlighted the advantages of verbalization as a complementary medium to visual explanations \cite{sevastjanova-2018-going, hohman-2019-telegam, szymanski-2021-visual, chromik-2021-shaprap}. \citet{zhang-lim-2022-relatable-xai} advocated for adding concepts and associations to make explanations more understandable, particularly in contrastive setups.

\citet{hsu-tan-2021-decision} introduced the task of \textbf{decision-focused summarization}.
While there are overlaps in the selection of important subsets of the input,
the textual nature of the output and the employment of saliency methods, 
our work is concerned with summarizing the token-level information provided by a saliency map from an arbitrary source for a single instance.
\citet{okeson-2021-summarize} found in their study that global feature attributions obtained by ranking features by different summary statistics helped users to communicate what the model had learned and to identify next steps for debugging it.
\citet{ronnqvist-2022-explaining} aggregated attribution scores from multiple documents to find top-ranked keywords for classes.

\begin{table}[t!]
    \centering
    
    \resizebox{\columnwidth}{!}{%
        \renewcommand{\arraystretch}{0.8}
    
        \begin{tabular}{r|rr}
        \toprule
        & \textsf{IMDb} & \textsf{AG News} \\        
        \midrule 
        \textbf{Saliency-related} & 100.00 & 99.17 \\
        \textit{``because of the high importance scores } \\ 
        \textit{of words such as 'oil', 'supply', [...]''} \\
    
        \textbf{Correct interpretation of neg. saliency} & 72.50 & 100.00 \\
        \textit{``[...] predicted this movie review as} \\ 
        \textit{'negative sentiment' because of the} \\
        \textit{high negative importance scores [...]''} \\
        
        \textbf{Suspecting a wrong prediction} & 55.00 & 23.21 \\
        \textit{``[...] it is unclear why the classifier} & FP: 0.00 & FP: 0.83 \\ 
        \textit{predicted this article as 'Business'.''} \\
    
        \midrule  
        
        \textbf{Associations} & 47.50 & 90.00 \\ 
        \textit{``These words are associated with} \\
        \textit{positive emotions and experiences.''} \\
        
        \textbf{Summarizations} & 10.00 & 27.50 \\
        \textit{``[...] the reviewer enjoyed these} \\
        \textit{aspects of the movie.''} \\
    
        \bottomrule
        \end{tabular}
    }
    \caption{Occurrences of semantics and accuracies of task comprehension (both in \%) in GPT-3.5-generated verbalizations for both datasets. FP = False positives.
    }
    \label{tab:gptquality}
\end{table}

In early explainability literature, \citet{van-lent-2004-explainable} already used \textbf{template filling}.
Templates in NLE frameworks were engineered by \citet{camburu-2020-make} 
to find inconsistencies in generated explanations.
While their templates were designed to mimic commonsense logic patterns present in the \textsf{e-SNLI} dataset \cite{camburu-2018-esnli},
our templates are a means to verbalize arbitrary saliency maps.
\citet{paranjape-2021-prompting} crafted templates
and used a mask-infilling approach to produce contrastive explanations from pre-trained language models.
\citet{donadello-dragoni-2021-bridging} utilized a template system to render explanation graph structures as text.
Recently, \citet{tursun-2023-self-explainability} used templates together with \lm{ChatGPT} prompts to generate captions containing verbalized saliency map explanations in the computer vision domain. However, they did not conduct an automated or human evaluation.

\section{Conclusion}
We conducted a comparative study on explanation representations.
We formalized the task of translating feature attributions into natural language and proposed two kinds of saliency map verbalization methods.
Instruction-based verbalizations outperformed all other saliency map representations on human ratings, indicating their summarization and contextualization capabilities are a necessary component in making saliency maps more accessible to humans, but they are still unreliable in terms of ensuring faithfulness and are dependant on a closed-source black-box model.
We find that template-based saliency map verbalizations reduce the cognitive load for humans
and are a viable option to improve on the ease of understanding of heatmaps without the need for additional resources.

\section*{Limitations}
\label{sec:limitations}

Our experimental setup excludes free-text rationales explaining the decisions of a model \cite{wiegreffe-2022-reframing,camburu-2018-esnli}, because their output is not based on attribution scores or highlighted spans of the input text, so we argue that they are not trivially comparable. However, there are end-to-end rationalization frameworks that can accommodate arbitrary saliency methods \cite{jain-2020-fresh,chrysostomou-aletras-2021-enjoy,ismail-2021-improving,atanasova-2022-diagnostics-guided,majumder-2022-rexc}, but require large language models that are expensive to train and perform inference with, so this is out of scope for this study. However, we also see that high-quality free-text rationales can be more easily generated with LLMs \cite{wang-2023-pinto,ho-2023-wikiwhy}, and a comparison between them and our attribution-based explanations is an interesting avenue for future work.

Inferring high-quality explanations from large language models necessitates excessive amounts of compute and storage. Although \lm{GPT} verbalizations are most promising, we urge the research community to look into more efficient ways to achieve similar results. In the future, we will explore if training a smaller model on top of the collected rationale-augmented verbalizations is feasible.

Emphasizing the concerns of \citet{rogers-2023-closed-ai}, we do not recommend the black-box model \lm{GPT-3.5} as a baseline for interpretability, because the model's training data or internal parameters can not be accessed and the dangers of deprecation as well as the lack of reproducibility are serious concerns. However, we do think it has revealed great potential as a surface realization and contextualization tool for the task of saliency map verbalization.

The causality problem explained in \citet{jacovi-2023-diagnosing} is not solved by our verbalizations, as it is an inherent problem with feature attribution and rationalization. Future work includes verbalizations alongside counterfactuals, e.g. in interactive setups \cite{feldhus-2022-mediators,shen-2023-convxai}.

Although multiple models and explanation-generating methods are available, we specifically focus on one pair for both datasets (\lm{BERT} and Integrated Gradients), because the focus of our investigation is on the quality of the representation rather than the model.

Finally, explicitly modelling expected highlights to mitigate misalignments as reported on in \citet{schuff-2022-human}, \citet{jacovi-2023-neighboring} and \citet{prasad-2021-extent} is still unexplored.

\section*{Acknowledgments}

We would like to thank Aleksandra Gabryszak for her help setting up and conducting the human evaluation.
We are indebted to Elif Kara, Ajay Madhavan Ravichandran, Akhyar Ahmed, Tatjana Zeen, Shushen Manakhimova, Raia Abu Ahmad, Ekaterina Borisova, Jo\~{a}o Lucas Mendes de Lemos Lins, Qianli Wang, and Sahil Chopra for their valuable work as annotators and evaluators.
We thank Gabriele Sarti, Arne Binder, Yuxuan Chen, and David Harbecke as well as the reviewers of the NLRSE Workshop at ACL 2023 for their feedback on the paper. 
This work has been supported by the German Federal Ministry of Education and Research as part of the project XAINES (01IW20005).

\section*{Contributions}
\begin{itemize}[noitemsep,topsep=0pt,leftmargin=*]
    \item NF: Writing, implementation of baselines and summarized template-based verbalization, prompt design and data generation, evaluation of user study, illustrations.
    \item LH: Writing and supervision.
    \item MDN: Conception, implementation and illustration of search and scoring methods for template-based verbalization.
    \item CE: Data curation, implementation of instruction-based verbalization pipeline, validation and empirical results on search and scoring methods.
    \item RS: Initial idea and outline.
    \item SM: Supervision and funding.
\end{itemize}

\bibliography{anthology,custom}
\bibliographystyle{acl_natbib}

\appendix

\section{Token ranks}
\label{app:tokenranks}

Figures \ref{fig:tokenrank_imdb} and \ref{fig:tokenrank_ag} show the coverage of the verbalizations, which makes up one aspect of explainer-faithfulness.

\begin{figure}[!ht]
    \centering
    \resizebox{\columnwidth}{!}{%
        \begin{tikzpicture}
        \begin{axis}[
            title={\textsf{IMDb}},
            xlabel={k},
            ylabel={},
            xmin=1, xmax=10,
            ymin=0, ymax=80,
            xtick={1,2,3,4,5,6,7,8,9,10},
            ytick={0,20,40,60,80},
            legend style={at={(0.55,0.67)}, anchor=west},
            ymajorgrids=true,
            grid style=dashed,
        ]

        \addplot+[
            smooth,
            mark=none,
            color=gray,
            fill=gray,
            fill opacity=0.15,
        ] coordinates {
            (1,80) (2,80) (3,80) (4,80) (5,80) (6,80) (7,80) (8,80) (9,80) (10,80)
        } \closedcycle;
        \addlegendentry{Top $k$ tokens}
        
        \addplot+[
            smooth,
            mark=none,
            color=smvcolor,
            fill=smvcolor,
            fill opacity=0.5,
        ] coordinates {
            (1,70) (2,52) (3,37) (4,31) (5,26) (6,14) (7,18) (8,23) (9,18) (10,11)
        } \closedcycle;
        \addlegendentry{SMV$_{\text{Templ}}$}

        \addplot+[
            smooth,
            mark=none,
            color=gptcolor,
            fill=gptcolor,
            fill opacity=0.5,
        ] coordinates {
            (1,53) (2,43) (3,29) (4,33) (5,26) (6,24) (7,18) (8,24) (9,21) (10,21)
        } \closedcycle;
        \addlegendentry{SMV$_{\texttt{GPT}}$}
        
        \end{axis}
        \end{tikzpicture}
    }
    \caption{Number of SMVs mentioning top $k$ attributed tokens in \textsf{IMDb}.}
    \label{fig:tokenrank_imdb}
\end{figure}

\begin{figure}[!ht]
    \centering
    \resizebox{\columnwidth}{!}{%
        \begin{tikzpicture}
        \begin{axis}[
            title={\textsf{AG News}},
            xlabel={k},
            ylabel={},
            xmin=1, xmax=10,
            ymin=0, ymax=120,
            xtick={1,2,3,4,5,6,7,8,9,10},
            ytick={0,20,40,60,80,100,120},
            legend style={at={(0.55,0.67)}, anchor=west},
            ymajorgrids=true,
            grid style=dashed,
        ]

        \addplot+[
            smooth,
            mark=none,
            color=gray,
            fill=gray,
            fill opacity=0.15,
        ] coordinates {
            (1,120) (2,120) (3,120) (4,120) (5,120) (6,120) (7,120) (8,119) (9,117) (10,114)
        } \closedcycle;
        \addlegendentry{Top $k$ tokens}
        
        \addplot+[
            smooth,
            mark=none,
            color=smvcolor,
            fill=smvcolor,
            fill opacity=0.5,
        ] coordinates {
            (1,97) (2,92) (3,89) (4,78) (5,50) (6,32) (7,6) (8,15) (9,7) (10,11)
        } \closedcycle;
        \addlegendentry{SMV$_{\text{Templ}}$}

        \addplot+[
            smooth,
            mark=none,
            color=gptcolor,
            fill=gptcolor,
            fill opacity=0.5,
        ] coordinates {
            (1,77) (2,51) (3,41) (4,33) (5,23) (6,22) (7,18) (8,16) (9,19) (10,18)
        } \closedcycle;
        \addlegendentry{SMV$_{\texttt{GPT}}$}
        
        \end{axis}
        \end{tikzpicture}
    }
    \caption{Number of SMVs mentioning top $k$ attributed tokens in \textsf{AG News}.}
    \label{fig:tokenrank_ag}
\end{figure}

\section{Automated simulatability evaluation}
\label{app:autosim}
\begin{table}[t!]
    \centering
    
    \resizebox{\columnwidth}{!}{%
        \renewcommand{\arraystretch}{0.8}
    
        \begin{tabular}{r|cccc}
        \toprule
        & \multicolumn{2}{c}{\textsf{AG News}} & \multicolumn{2}{c}{\textsf{IMDb}} \\
        & $S_{V}$ & $W + S_{V}$ & $S_{V}$ & $W + S_{V}$ \\
        
        \midrule
        
        Conv. Search 
            & 91.73 & 94.10 & 86.08 & \textbf{96.00} \\
        Span Search
            & 87.16 & \textbf{94.39} & 89.08 & 95.90 \\
        Top $k=5$ tokens
            & \textbf{92.54} & 93.93 & 92.38 & 95.60 \\
        SMV$_{\text{Templ}}$ 
            & 91.94 & 94.10 & \textbf{94.26} & 94.90 \\
        SMV$_{\text{\lm{GPT}}}$ 
            & 69.16 & 70.00 & 81.25 & 81.25 \\
    
        \bottomrule
        \end{tabular}
    }
    \caption{Automated simulatability evaluation (Accuracy in \%) using a \lm{T5-large} model (Accuracy on original input: \textsf{AG News} – 92.58\%; \textsf{IMDb} – 97.62\%) to reproduce the underlying BERT model's prediction based on only seeing one of the verbalizations $S_V$ (prepended by the original input $W$).
    }
    \label{tab:autosim}
\end{table}

We follow \citet{wiegreffe-2021-measuring} and \citet{hase-2020-leakage} and train a second language model to simulate the behavior of the explained BERT model.
Table~\ref{tab:autosim} shows the simulation accuracy of a \lm{T5-large} receiving various types of verbalizations (plus the original input). We can see for both datasets that SMV$_\text{\lm{GPT}}$ induces the most noise and thus results in the lowest accuracy, while the raw output of the search methods (Conv/Span) are most faithful in combination with the original input.

\section{Efficiency}
\label{app:efficiency}
First, we measure a runtime of less than two minutes on a CPU (i5-12600k) to generate template-based verbalizations for all 25k instances of \textsf{IMDb}.
Given pre-computed saliency maps from any explainer, this is considerably faster than using an end-to-end model for extractive rationales, e.g. \citet{treviso-martins-2020-explanation}, which takes several hours for training and then more than 10 minutes for inference on an RTX 3080 GPU.
\lm{GPT-3.5} with at least 175B parameters, which obliterates the other two setups. This means that there is a considerable carbon footprint associated with using it. Future work has to look into training considerably smaller models on the generated verbalizations.

\begin{figure*}[ht!]
    \centering
    \resizebox{\textwidth}{!}{%
        \includegraphics{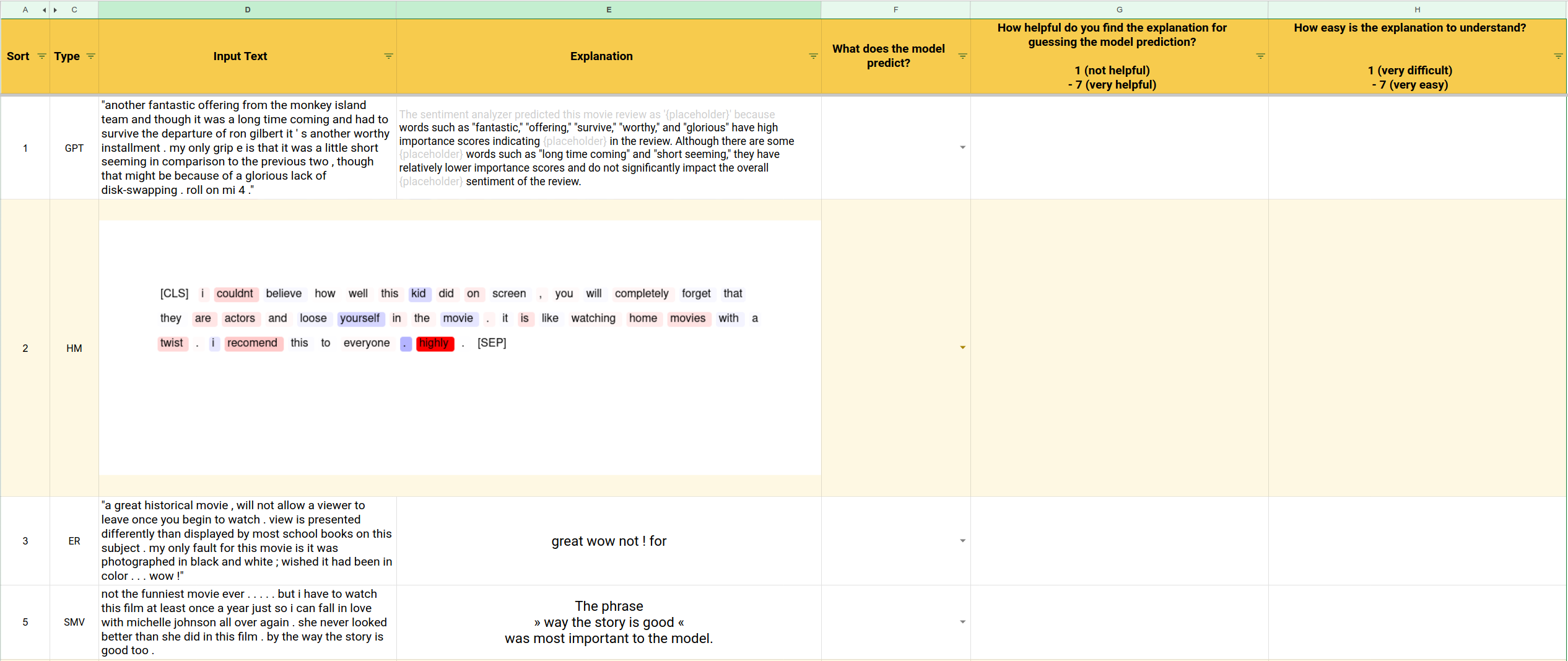}
    }
    \resizebox{\textwidth}{!}{%
        \includegraphics{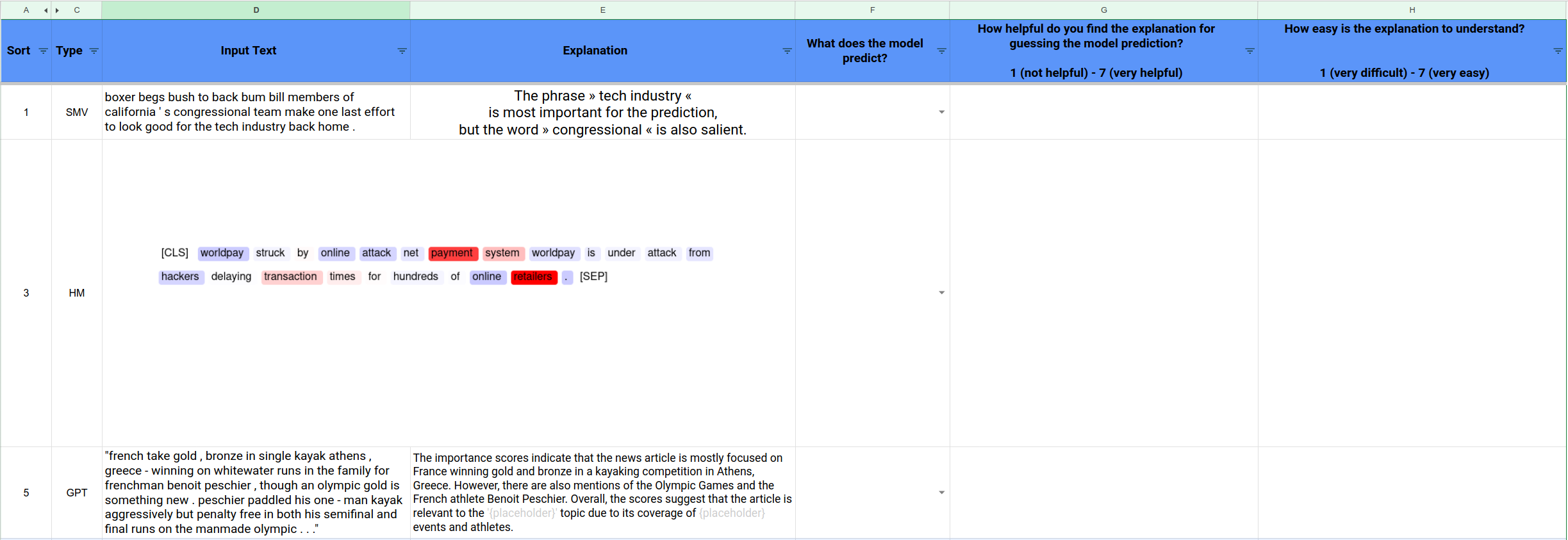}
    }
    \caption{Annotation spreadsheet including one instance from every type of explanation representation in \textsf{IMDb} and \textsf{AG News}, as used in the human evaluation described in \S \ref{sec:human_eval}.}
    \label{fig:interface}
\end{figure*}

\section{Subset selection heuristics}
\label{app:heuristics}
\begin{itemize}[noitemsep,topsep=0pt,leftmargin=*]    
    \item We restrict our experiments to explaining a single outcome -- the predicted label $\hat{y}$ -- and thus modify our metric (Eq.~\ref{eqn:cov}): $\text{Cov}_{+}$ only considers the positive attributions $s_i > 0$.
    
    \item We select instances achieving at least a $\text{Cov}_{+}$ score of 15\% (indicating the attribution mass is not too evenly distributed, making interpretations of saliency maps challenging).

    \item We find values for $q$ (\S \ref{sec:compare}) of $0.5 \leq q \leq 0.75$ to produce the right amount of candidates in the end, s.t. there almost always is at least one candidate in the $q$-th quantile and the resulting verbalization is not longer than most text inputs.
    
    \item We only consider instances with a maximum token length of 80, s.t.\ the human evaluation is more manageable for annotators.
    
    \item We select equal amounts of instances for every true label $y$ (\textsf{IMDb}: 40 positive + 40 negative sentiment; \textsf{AG News}: 30 World + 30 Sports + 30 Business + 30 Sci/Tech) in each dataset.

    \item We select 25\% of \textsf{IMDb} and 46.67 \% of \textsf{AG News} to be false predictions by the \lm{BERT} model ($y \neq \hat{y}$).
\end{itemize}

We apply the weighted average for \textsf{IMDb-BERT-IG} ($\beta$ = 0.4) and the quantile scoring metric for \textsf{AG News-BERT-IG} ($n = 3$).
We chose the number of candidates to be $k=5$ in all cases and the threshold $q$ to be $.75$ for \textsf{IMDb} and \textsf{AG News} as the average length of the input is lower for the latter which results in too few candidates with higher $q$s.

\section{Templates for Verbalizing Explanations}
\label{app:templates}

\begin{figure*}[t!]

    \centering
    \resizebox{\textwidth}{!}{%
    \begin{tabular}{rr}
    
    \toprule
    \multicolumn{2}{c}{\textbf{Examples for leading sentence}} \\
    The words \{$w_1$\}, \{$\dots$\}, and \{$w_n$\} are most important. & Most important is \{$\dots$\} \\
    The most salient features are \{$\dots$\} & The model predicted this label, because \{$\dots$\} \\
    $\dots$ is the span that was most important. \\

    \midrule 
    \textbf{Features or linguistic units} & \textbf{More than one unit} \\
    feature(s) & The two phrases \{$\dots$\} and \{$\dots$\} \\
    word(s) & Both phrases \{$\dots$\} and \{$\dots$\} \\
    token(s) & $\dots$ are both salient. \\
    phrase(s) & The (top) three most important tokens $\dots$ \\ 
    punctuation & $\dots$ words such as \{$\dots$\} and \{$\dots$\} \\

    \midrule 
    \textbf{Synonyms for \textit{important}} & \textbf{Conjunctions \& Adverbs} \\ 
    salient & \{$\dots$\}, while \{$\dots$\} \\
    influential & \{$\dots$\}, whereas \{$\dots$\} \\
    key & $\dots$ also salient \\
    impactful & with the word \{$\dots$\} also being salient. \\

    \midrule 
    \textbf{Additions for \textit{important} \{$\dots$\}} & \textbf{Variations of \textit{important}} \\
    $\dots$ for (the/this) prediction. & $\dots$ focused on the most for this prediction. \\
    $\dots$ (to the model) in (making/predicting & $\dots$ used by the model to make its prediction. \\
    choosing/producing/shaping) this outcome. & $\dots$ caused the model to predict this outcome. \\
    $\dots$ with respect to the outcome. & indicate the model's predicted label. \\
    $\dots$ in this text. & $\dots$ shaped the model's outcome (the most). \\

    \midrule 
    \textbf{Synonyms for \textit{prediction}} & \textbf{Polarity} \\ 
    outcome & \{$\dots$\} is least important. \\
    model('s) prediction & \{$\dots$\} is more salient than \{$\dots$\}. \\
    model's judgment & \{$\dots$\} is less influential than \{$\dots$\}. \\ 
    model('s) behavior & \\
    prediction of the classifier \\
    (model's) predicted label \\
    decision \\

    \midrule
    \end{tabular}
    }

    \centering
    \resizebox{\textwidth}{!}{%
    \begin{tabular}{rr}

    \multicolumn{2}{c}{\textbf{Dataset-specific}} \\
    \textsf{IMDb} & \textsf{AG News} \\ 
    \{$\dots$\} for the sentiment label. 
        & \{$\dots$\} indicative of the model's topic classification. \\ 
    \{$\dots$\} most indicative of the sentiment.
        & \{$\dots$\} in this article. \\ 
    \{$\dots$\} most indicative for the sentiment analysis.
        & The most salient words in this article are \{$\dots$\}. \\
    \{$\dots$\} used by the model to predict this sentiment label.
        & \{$\dots$\}, because \{$\dots$\} appeared in the article. \\

    \bottomrule
    \end{tabular}
    }

    \caption{Templates for model-free saliency map verbalization.
    }
    \label{tab:templates}

\end{figure*}

We design our templates as atomic expressions with constraints and blanks that can be filled with words from $W$.
In the most basic cases, we refer to spans, phrases, words and characters as salient or important for some prediction.
We design the templates to express saliency information concisely and enable users to reproduce the model's decision process (simulatability).
The set of templates is depicted in Table~\ref{tab:templates}.

Our template-based methodology is task- and model-invariant by design, because no task-specific model or NLG component is involved.
Achieving sufficiency (measured by coverage) is harder, because a full translation of any saliency map is too verbose and thus not helpful.

\section{List of LLM prompts}
\label{app:prompts}

At first, we treated this as table-to-text task -- which has recently been tackled with prompt-based large language models \cite{chen-2023-table-reasoners,xiang-2022-asdot} -- where we provided a list of attribution scores and, separate from that, a list of tokens. However, we registered less hallucinations (the model incorrectly mapping between words and their scores) when we provided the input as a joint representation as shown in Fig.~\ref{fig:gptmethod}.

For the two datasets, we then used the token+score representation as \texttt{\string{sample\string}} and a \texttt{\string{label\_str\string}} being the predicted label (\textsf{IMDb}: \textit{positive} or \textit{negative}; \textsf{AG News}: \textit{Worlds}, \textit{Sports}, \textit{Business}, or \textit{Sci/Tech}) and wrote the instructions in Fig.~\ref{fig:gpt-instructions}.

\begin{figure*}[t!]
    \centering
        
        \begin{displayquote}
        \textbf{\textsf{IMDb}}
        Movie review with importance scores: \texttt{\string{sample\string}}. \\
        A sentiment analyzer has predicted this text as '\texttt{\string{label\_str\string}} sentiment'. The scores behind each word indicate how important it was for the analyzer to predict '\texttt{\string{label\_str\string}} sentiment'. The scores have been determined after the sentiment analyzer has already made its prediction. The sentiment analyzer cannot base its prediction on the scores, only on the movie review itself. \\
        Based on the importance scores, briefly explain why the sentiment analyzer has predicted this movie review as '\texttt{\string{label\_str\string}} sentiment': 
        \end{displayquote}

        \begin{displayquote}
        \textbf{\textsf{AG News} (Figure \ref{fig:heatmap}, r.)} \\
        News article with importance scores: \texttt{\string{sample\string}}. \\
        A topic classifier has predicted this text as '\texttt{\string{label\_str\string}}'. The scores behind each word indicate how important it was for the classifier to predict '\texttt{\string{label\_str\string}}'. The scores have been determined after the topic classifier has already made its prediction. The topic classifier cannot base its prediction on the scores, only on the news article itself. \\
        Based on the importance scores, briefly explain why the topic classifier has predicted this news article as '\texttt{\string{label\_str\string}}': 
        \end{displayquote}
        
    \caption{Task instructions applied to \textsf{IMDb} and \textsf{AG News} used by \lm{GPT-3.5} (see App.~\ref{app:prompts} for details).}
    \label{fig:gpt-instructions}

\end{figure*}

\section{Post-processing of GPT outputs}
\label{app:gpt-postproc}

\begin{table*}[ht]
    \centering
    \resizebox{\textwidth}{!}{%
    \begin{tabular}{rrrrr}
        \textsf{IMDb} & \multicolumn{4}{c}{\textsf{AG News}} \\

        \textbf{Classes}
            & \textbf{Sports}
            & \textbf{Business}
            & \textbf{World}
            & \textbf{Sci/Tech} \\
        \toprule
        
        positivity (+)
            & sport
            & businesses
            & global
            & science
            \\
        negativity (-)
            & the world of sports
            & business and economics
            & global politics
            & science and technology
            \\

            & 
            & business and finance
            & international
            & scientific
            \\

            & 
            & economics
            & all over the world
            & tech
            \\

            & 
            & finance
            & global issues
            & technical
            \\

            & 
            & financial
            & global affairs
            & technology
            \\

            & 
            & the business world
            & international relations
            & technological
            \\

            & 
            & the economy
            & a global issue or event
            & the tech industry
            \\

            & 
            & corporate finance
            & 
            & the technology industry
            \\
        
        \bottomrule
        
    \end{tabular}
    }
    \caption{Post-processing of \lm{GPT-3.5} verbalizations for human evaluation.}
    \label{tab:postproc}
\end{table*}

\paragraph*{\textsf{AG News}}

In order to prevent label leakage, we employed the string replacements listed in Tab.~\ref{tab:postproc}. In our human evaluation, they were replaced with "$\{$placeholder$\}$", so annotators could perform the simulatability task without cheating.

\end{document}